# Towards Understanding and Harnessing the Potential of Clause Learning

**Paul Beame**                                                                beame@cs.washington.edu
**Henry Kautz**                                                               kautz@cs.washington.edu
**Ashish Sabharwal**                                                          ashish@cs.washington.edu
*Computer Science and Engineering*
*University of Washington, Box 352350*
*Seattle, WA 98195-2350, USA*

## Abstract

Efficient implementations of DPLL with the addition of clause learning are the fastest complete Boolean satisfiability solvers and can handle many significant real-world problems, such as verification, planning and design. Despite its importance, little is known of the ultimate strengths and limitations of the technique. This paper presents the first precise characterization of clause learning as a proof system (CL), and begins the task of understanding its power by relating it to the well-studied resolution proof system. In particular, we show that with a new learning scheme, CL can provide exponentially shorter proofs than many proper refinements of general resolution (RES) satisfying a natural property. These include regular and Davis-Putnam resolution, which are already known to be much stronger than ordinary DPLL. We also show that a slight variant of CL with unlimited restarts is as powerful as RES itself. Translating these analytical results to practice, however, presents a challenge because of the nondeterministic nature of clause learning algorithms. We propose a novel way of exploiting the underlying problem structure, in the form of a high level problem description such as a graph or PDDL specification, to guide clause learning algorithms toward faster solutions. We show that this leads to exponential speed-ups on grid and randomized pebbling problems, as well as substantial improvements on certain ordering formulas.

## 1. Introduction

In recent years the task of deciding whether or not a given CNF propositional logic formula is satisfiable has gone from a problem of theoretical interest to a practical approach for solving real-world problems. Satisfiability (SAT) procedures are now a standard tool for hardware verification, including verification of super-scalar processors (Velev & Bryant, 2001; Biere et al., 1999a). Open problems in group theory have been encoded and solved using satisfiability solvers (Zhang & Hsiang, 1994). Other applications of SAT include circuit diagnosis and experiment design (Konuk & Larrabee, 1993; Gomes et al., 1998b).

The most surprising aspect of such relatively recent practical progress is that the best complete satisfiability testing algorithms remain variants of the Davis-Putnam-Logemann-Loveland or DPLL procedure (Davis & Putnam, 1960; Davis et al., 1962) for backtrack search in the space of partial truth assignments. The key idea behind its efficacy is the pruning of search space based on falsified clauses. Since its introduction in the early 1960's, the main improvements to DPLL have been smart branch selection heuristics (*e.g.,* Li &

Anbulagan, 1997), and extensions such as randomized restarts (Gomes et al., 1998a) and clause learning (see *e.g.,* Marques-Silva & Sakallah, 1996). One can argue that of these, clause learning has been the most significant in scaling DPLL to realistic problems. This paper attempts to understand the potential of clause learning and suggests ways to harness its power.

Clause learning grew out of work in AI on explanation-based learning (EBL), which sought to improve the performance of backtrack search algorithms by generating explanations for failure (backtrack) points, and then adding the explanations as new constraints on the original problem (de Kleer & Williams, 1987; Stallman & Sussman, 1977; Genesereth, 1984; Davis, 1984). For general constraint satisfaction problems the explanations are called "conflicts" or "no goods"; in the case of Boolean CNF satisfiability, the technique becomes clause learning – the reason for failure is learned in the form of a "conflict clause" which is added to the set of given clauses. A series of researchers (Bayardo Jr. & Schrag, 1997; Marques-Silva & Sakallah, 1996; Zhang, 1997; Moskewicz et al., 2001; Zhang et al., 2001) showed that clause learning can be efficiently implemented and used to solve hard problems that cannot be approached by any other technique.

Despite its importance there has been little work on formal properties of clause learning, with the goal of understanding its fundamental strengths and limitations. A likely reason for such inattention is that clause learning is a rather complex rule of inference – in fact, as we describe below, a complex family of rules of inference. A contribution of this paper is a precise mathematical specification of various concepts used in describing clause learning.

Another problem in characterizing clause learning is defining a formal notion of the strength or power of a reasoning method. This paper uses the notion of proof complexity (Cook & Reckhow, 1977), which compares inference systems in terms of the sizes of the shortest proofs they sanction. We use CL to denote clause learning viewed as a proof system. A family of formulas $C$ provides an *exponential separation* between systems $S_1$ and $S_2$ if the shortest proofs of formulas in $C$ in system $S_1$ are exponentially smaller than the corresponding shortest proofs in $S_2$. From this basic propositional proof complexity point of view, only families of unsatisfiable formulas are of interest, because only proofs of unsatisfiability can be large; minimum proofs of satisfiability are linear in the number of variables of the formula. Nevertheless, Achlioptas et al. (2001) have shown how negative proof complexity results for unsatisfiable formulas can be used to derive time lower bounds for specific inference algorithms running on satisfiable formulas as well.

Proof complexity does not capture everything we intuitively mean by the power of a reasoning system, because it says nothing about how difficult it is to find shortest proofs. However, it is a good notion with which to begin our analysis, because the size of proofs provides a lower bound on the running time of any implementation of the system. In the systems we consider, a branching function, which determines which variable to split upon or which pair of clauses to resolve, guides the search. A negative proof complexity result for a system tells us that a family of formulas is intractable even with a perfect branching function; likewise, a positive result gives us hope of finding a branching function.

A basic result in proof complexity is that general resolution, denoted RES, is exponentially stronger than the DPLL procedure (Bonet et al., 2000; Ben-Sasson et al., 2000). This is because the trace of DPLL running on an unsatisfiable formula can be converted to a *tree-like* resolution proof of the same size, and tree-like proofs must sometimes be exponentially larger than the DAG-like proofs generated by RES. Although RES can yield shorter proofs, in practice DPLL is better because it provides a more efficient way to search for proofs. The weakness of the tree-like proofs that DPLL generates is that they do not reuse derived clauses. The conflict clauses found when DPLL is augmented by clause learning correspond to reuse of derived clauses in the associated resolution proofs and thus to more general forms of resolution proofs. As a theoretical upper bound, all DPLL based approaches, including those involving clause learning, are captured by RES. An intuition behind the results in this paper is that the addition of clause learning moves DPLL closer to RES while retaining its practical efficiency.

It has been previously observed that clause learning can be viewed as adding resolvents to a tree-like proof (Marques-Silva, 1998). However, this paper provides the first mathematical proof that clause learning, viewed as a propositional proof system CL, is exponentially stronger than tree-like resolution. This explains, formally, the performance gains observed empirically when clause learning is added to DPLL based solvers. Further, we describe a generic way of extending families of formulas to obtain ones that exponentially separate CL from many refinements of resolution known to be intermediate in strength between RES and tree-like resolution. These include *regular* and *Davis-Putnam* resolution, and any other proper refinement of RES that behaves naturally under restrictions of variables, *i.e.*, for any formula $F$ and restriction $\rho$ on its variables, the shortest proof of $F|_\rho$ in the system is not any larger than a proof of $F$ itself. The argument used to prove this result involves a new clause learning scheme called FirstNewCut that we introduce specifically for this purpose. Our second technical result shows that combining a slight variant of CL, denoted CL--, with unlimited restarts results in a proof system as strong as RES itself. This intuitively explains the speed-ups obtained empirically when randomized restarts are added to DPLL based solvers, with or without clause learning.

Given these results about the strengths and limitations of clause learning, it is natural to ask how the understanding we gain through this kind of analysis may lead to practical improvement in SAT solvers. The theoretical bounds tell us the potential power of clause learning; they don't give us a way of *finding* short solutions when they exist. In order to leverage their strength, clause learning algorithms must follow the "right" variable order for their branching decisions for the underlying DPLL procedure. While a good variable order may result in a polynomial time solution, a bad one can make the process as slow as basic DPLL without learning. The latter half of this paper addresses this problem of moving from analytical results to practical improvement. The approach we take is the use of the problem structure for guiding SAT solvers in their branch decisions.

Both random CNF formulas and those encoding various real-world problems are quite hard for current SAT solvers. However, while DPLL based algorithms with lookahead but no learning (such as `satz` by Li & Anbulagan, 1997) and those that try only one carefully chosen assignment without any backtracks (such as `SurveyProp` by Mézard & Zecchina, 2002) are our best tools for solving random formula instances, formulas arising from various real applications seem to require clause learning as a critical ingredient. The key thing that makes this second class of formulas different is the inherent structure, such as dependence graphs in scheduling problems, causes and effects in planning, and algebraic structure in group theory.

Most theoretical and practical problem instances of satisfiability problems originate, not surprisingly, from a higher level description, such as planning domain definition language or PDDL specification for planning, timed automata or logic description for model checking, task dependency graph for scheduling, circuit description for VLSI, algebraic structure for group theory, and processor specification for hardware. Typically, this description contains more structure of the original problem than is visible in the flat CNF representation in DIMACS format (Johnson & Trick, 1996) to which it is converted before being fed into a SAT solver. This structure can potentially be used to gain efficiency in the solution process. While there has been work on extracting structure after conversion into a CNF formula by exploiting variable dependency (Giunchiglia et al., 2002; Ostrowski et al., 2002), constraint redundancy (Ostrowski et al., 2002), symmetry (Aloul et al., 2002), binary clauses (Brafman, 2001), and partitioning (Amir & McIlraith, 2000), using the original higher level description itself to generate structural information is likely to be more effective. The latter approach, despite its intuitive appeal, remains largely unexplored, except for suggested use in bounded model checking (Shtrichman, 2000) and the separate consideration of cause variables and effect variables in planning (Kautz & Selman, 1996).

In this paper, we further open this line of research by proposing an effective method for exploiting problem structure to guide the branching decision process of clause learning algorithms. Our approach uses the original high level problem description to generate not only a CNF encoding but also a *branching sequence* that guides the SAT solver toward an efficient solution. This branching sequence serves as auxiliary structural information that was possibly lost in the process of encoding the problem as a CNF formula. It makes clause learning algorithms learn useful clauses instead of wasting time learning those that may not be reused in future at all. We give an exact sequence generation algorithm for pebbling formulas, using the underlying pebbling graph as the high level description. We also give a much simpler but approximate branching sequence generation algorithm for $GT_n$ formulas, utilizing their underlying ordering structure. Our sequence generators work for the 1UIP learning scheme (Zhang et al., 2001), which is one of the best known. They can also be extended to other schemes, including FirstNewCut. Our empirical results are based on our extension of the popular SAT solver `zChaff` (Moskewicz et al., 2001).

We show that the use of branching sequences produced by our generator leads to exponential empirical speedups for the class of grid and randomized pebbling formulas. These formulas, more commonly occurring in theoretical proof complexity literature (Ben-Sasson et al., 2000; Beame et al., 2003a), can be thought of as representing precedence graphs in dependent task systems and scheduling scenarios. They can also be viewed as restricted planning problems. Although admitting a polynomial size solution, both grid and randomized pebbling problems are not so easy to solve deterministically, as is indicated by our experimental results for unmodified `zChaff`. We also report significant gains obtained for the class of $GT_n$ formulas which, again, have appeared frequently in proof complexity results (Krishnamurthy, 1985; Bonet & Galesi, 2001; Alekhnovich et al., 2002). From a broader perspective, our results for pebbling and $GT_n$ formulas serve as a proof of concept that analysis of problem structure can be used to achieve dramatic improvements even in the current best clause learning based SAT solvers.

## 2. Preliminaries

A CNF formula $F$ is an AND ($\wedge$) of *clauses*, where each clause is an OR ($\vee$) of *literals*, and a literal is a variable or its negation ($\neg$). It is natural to think of $F$ as a set of clauses and each clause as a set of literals. A clause that is a subset of another is called its *subclause*. The *size* of $F$ is the number of clauses in $F$.

Let $\rho$ be a partial assignment to the variables of $F$. The *restricted formula* $F^\rho$ is obtained from $F$ by replacing variables in $\rho$ with their assigned values. $F$ is said to be *simplified* if all clauses with at least one TRUE literal are deleted and all occurrences of FALSE literals are removed from clauses. $F|_\rho$ denotes the result of simplifying the restricted formula $F^\rho$.

### 2.1 The DPLL Procedure

The basic idea of the Davis-Putnam-Logemann-Loveland (DPLL) procedure (Davis & Putnam, 1960; Davis et al., 1962) for testing satisfiability of CNF formulas is to *branch* on variables, setting them to TRUE or FALSE, until either an initial clause is *violated* (*i.e.* has all literals set to FALSE) or no more clauses remain in the simplified residual formula. In the former case, we backtrack to the last branching decision whose other branch has not been tried yet, reverse the decision, and proceed recursively. In the latter, we terminate with a satisfying assignment. If all possible branches have been unsuccessfully tried, the formula is declared unsatisfiable. To increase efficiency, *unit clauses* (those with only one unset literal) are immediately set to true. *Pure literals* (those whose negation does not appear) are also set to TRUE as a preprocessing step and, in some implementations, in the simplification process after every branch.

In this paper, we will use the term DPLL to denote the basic branching and backtracking procedure described above. It will not include learning conflict clauses when backtracking, but will allow intelligent branching heuristics as well as common extensions such as fast backtracking and restarts. Note that this is in contrast with the occasional use of the term DPLL to encompass practically all branching and backtracking approaches to SAT, including those involving learning.

### 2.2 Proof Systems

A *propositional proof system* (Cook & Reckhow, 1977) is a polynomial time computable predicate $S$ such that a propositional formula $F$ is unsatisfiable iff there exists a *proof* $p$ for which $S(F, p)$ holds. In other words, it is an efficient (in the size of the proof) procedure to check the correctness of proofs presented in a certain format. Finding short proofs, however, may still be difficult. In fact, short proofs may not exist in the proof system if it is too weak. In the rest of this paper, we refer to such systems simply as *proof systems* and omit the word propositional.

**Definition 1.** For a proof system $S$ and an unsatisfiable formula $F$, the *complexity* of $F$ in $S$, denoted $\mathcal{C}_S(F)$, is the length of the shortest refutation of $F$ in $S$. For a family $\{F_n\}$ of formulas over increasing number of variables $n$, its asymptotic complexity in $S$, denoted $\mathcal{C}_S(F_n)$ with abuse of notation, is measured with respect to the increasing sizes of $F_n$.

**Definition 2.** For proof systems $S$ and $T$, and a function $f : \mathbb{N} \to \mathbb{N}$,

- $S$ is *natural* if for any formula $F$ and restriction $\rho$ on its variables, $\mathcal{C}_S(F|_\rho) \leq \mathcal{C}_S(F)$.

- $S$ is a *refinement* of $T$ if proofs in $S$ are also (restricted) proofs in $T$.

- A refinement $S$ of $T$ is $f(n)$-*proper* if there exists a witnessing family $\{F_n\}$ of formulas such that $\mathcal{C}_S(F_n) \geq f(n) \cdot \mathcal{C}_T(F_n)$. The refinement is *exponentially-proper* if $f(n) = 2^{n^{\Omega(1)}}$ and *super-polynomially-proper* if $f(n) = n^{\omega(1)}$.

## 2.3 Resolution

Resolution (RES) is a widely studied simple proof system that can be used to prove unsatisfiability of CNF formulas. Our complexity results concerning the power of clause learning are in relation to this system. The *resolution rule* states that given clauses $(A \vee x)$ and $(B \vee \neg x)$, we can derive clause $(A \vee B)$ by *resolving on* $x$. A *resolution derivation* of $C$ from a CNF formula $F$ is a sequence $\pi = (C_1, C_2, \ldots, C_s \equiv C)$ where each clause $C_i$ is either a clause of $F$ (an *initial* clause) or derived by applying the resolution rule to $C_j$ and $C_k$, $j, k < i$ (a *derived* clause). The *size* of $\pi$ is $s$, the number of clauses occurring in it. We will assume that each $C_j \neq C$ in $\pi$ is used to derive at least one other clause $C_i, i > j$. Any derivation of the empty clause $\Lambda$ from $F$, also called a *refutation* or *proof* of $F$, shows that $F$ is unsatisfiable.

Despite its simplicity, unrestricted resolution is hard to implement efficiently due to the difficulty of finding good choices of clauses to resolve; natural choices typically yield huge storage requirements. Various restrictions on the structure of resolution proofs lead to less powerful but easier to implement refinements that have been studied well, such as tree-like, regular, linear, positive, negative, semantic, and Davis-Putnam resolution. *Tree-like resolution* uses non-empty derived clauses exactly once in the proof and is equivalent to an optimal DPLL procedure. *Regular resolution* allows any variable to be resolved upon at most once along any "path" in the proof from an initial clause to $\Lambda$, allowing (restricted) reuse of derived clauses. *Linear resolution* requires each clause $C_i$ in a derivation $(C_1, C_2, \ldots, C_s)$ to be either an initial clause or be derived by resolving $C_{i-1}$ with $C_j, j < i-1$. For any assignment $\alpha$ to the variables, an $\alpha$-*derivation* requires at least one clause involved in each resolution step to be falsified by $\alpha$. When $\alpha$ is the all FALSE assignment, the derivation is *positive*. When it is the all TRUE assignment, the derivation is *negative*. A derivation is *semantic* if it is an $\alpha$-derivation for some $\alpha$. *Davis-Putnam resolution*, also called *ordered resolution*, is a refinement of regular resolution where every sequence of variables in a path from an initial clause to $\Lambda$ respects the same ordering on the variables.

While all these refinements are sound and complete as proof systems, they differ in efficiency. For instance, regular, linear, positive, negative, semantic, and Davis-Putnam resolution are all known to be exponentially stronger than tree-like resolution (Bonet et al., 2000; Bonet & Galesi, 2001; Buresh-Oppenheim & Pitassi, 2003) whereas tree-like, regular, and Davis-Putnam resolution are known to be exponentially weaker than RES (Bonet et al., 2000; Alekhnovich et al., 2002).

**Proposition 1.** *Tree-like, regular, linear, positive, negative, semantic, and Davis-Putnam resolution are natural refinements of* RES.

**Proposition 2 (Bonet et al., 2000; Alekhnovich et al., 2002).** *Tree-like, regular, and Davis-Putnam resolution are exponentially-proper natural refinements of* RES.

### 2.4 Clause Learning

Clause learning (see *e.g.,* Marques-Silva & Sakallah, 1996) can be thought of as an extension of the DPLL procedure that caches causes of assignment failures in the form of learned clauses. It proceeds by following the normal branching process of DPLL until there is a "conflict" after unit propagation. If this conflict occurs when no variable is currently branched upon, the formula is declared unsatisfiable. Otherwise, the "conflict graph" is analyzed and the "cause" of the conflict is learned in the form of a "conflict clause." The procedure now backtracks and continues as in ordinary DPLL, treating the learned clause just like initial ones. A clause is said to be *known* at a stage if it is either an initial clause or has previously been learned.

The learning process is expected to save us from redoing the same computation when we later have an assignment that causes conflict due in part to the same reason. Variations of such conflict-driven learning include different ways of choosing the clause to learn (different *learning schemes*) and possibly allowing multiple clauses to be learned from a single conflict (Zhang et al., 2001). In the last decade, many algorithms based on this idea have been proposed and demonstrated to be empirically successful on large problems that could not be handled using other methodologies (Bayardo Jr. & Schrag, 1997; Marques-Silva & Sakallah, 1996; Zhang, 1997; Moskewicz et al., 2001). We leave a more detailed discussion of the concepts involved in clause learning as well as its formulation as a proof system CL to Section 3.

### 2.5 Pebbling Formulas

Pebbling formulas are unsatisfiable CNF formulas whose variations have been used repeatedly in proof complexity to obtain theoretical separation results between different proof systems (Ben-Sasson et al., 2000; Beame et al., 2003a). The version we will use in this paper is known to be easy for regular resolution but hard for tree-like resolution, and hence for DPLL without learning (Ben-Sasson et al., 2000). We use these formulas to show how one can utilize problem structure to allow clause learning algorithms to handle much bigger problems than they otherwise can.

Pebbling formulas represent the constraints for sequencing a system of tasks that need to be completed, where each task can be accomplished in a number of alternative ways. The associated pebbling graph has a node for each task, labeled by a disjunction of variables representing the different ways of completing the task. Placing a pebble on a node in the graph represents accomplishing the corresponding task. Directed edges between nodes denote task precedence; a node is pebbled when all of its predecessors in the graph are pebbled. The pebbling process is initialized by placing pebbles on all indegree zero nodes. This corresponds to completing those tasks that do not depend on any other.

Formally, a *Pebbling formula* $Pbl_G$ is an unsatisfiable CNF formula associated with a directed, acyclic *pebbling graph* $G$ (see Figure 1). Nodes of $G$ are labeled with disjunctions of variables, *i.e.* with clauses. A node labeled with clause $C$ is thought of as *pebbled* under a (partial) variable assignment $\sigma$ if $C|_\sigma = \text{TRUE}$. $Pbl_G$ contains three kinds of clauses – precedence clauses, source clauses and target clauses. For instance, a node labeled $(x_1 \vee x_2)$ with three predecessors labeled $(p_1 \vee p_2 \vee p_3)$, $q_1$ and $(r_1 \vee r_2)$ generates six precedence clauses $(\neg p_i \vee \neg q_j \vee \neg r_k \vee x_1 \vee x_2)$, where $i \in \{1,2,3\}, j \in \{1\}$ and $k \in \{1,2\}$. The precedence clauses imply that if all predecessors of a node are pebbled, then the node itself must also be pebbled. For every indegree zero *source node* $s$ of $G$, $Pbl_G$ contains the clause labeling $s$ as a source clause. Thus, $Pbl_G$ implies that all source nodes are pebbled. For every outdegree zero *target node* of $G$ labeled, say, $(t_1 \vee t_2)$, $Pbl_G$ has target clauses $\neg t_1$ and $\neg t_2$. These imply that target nodes are not pebbled, and provide a contradiction.

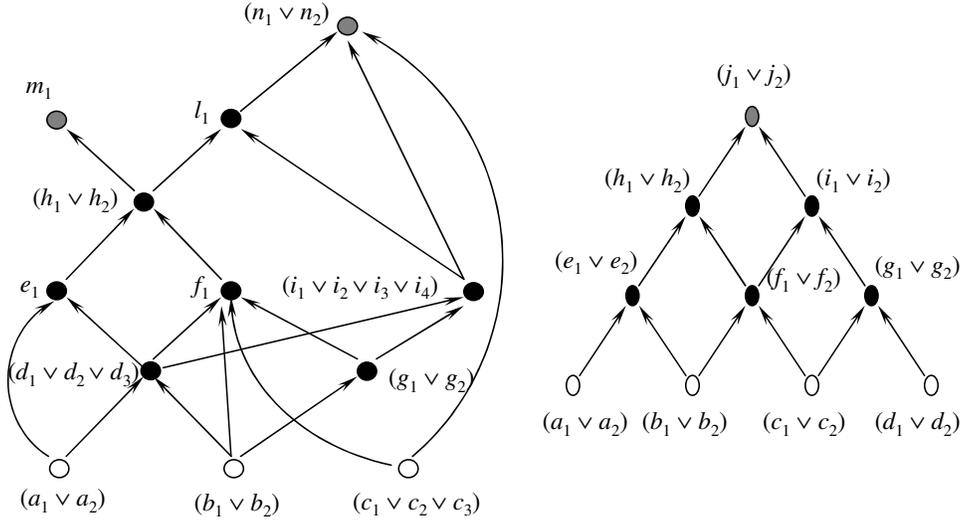

Figure 1: A general pebbling graph with distinct node labels, and a 4-layer grid pebbling graph

*Grid pebbling formulas* are based on simple pyramid-shaped layered pebbling graphs with distinct variable labels, 2 predecessors per node, and disjunctions of size 2 (see Figure 1). *Randomized pebbling formulas* are more complicated and correspond to random pebbling graphs. In this paper, we only consider pebbling graphs where no variable appears more than once in any node label. In general, random pebbling graphs allow multiple target nodes. However, the more the targets, the easier it is to produce a contradiction because we can focus only on the (relatively smaller) subgraph under the lowest target. Hence, for our experiments, we add a simple grid structure at the top of randomly generated pebbling formulas to make them have exactly one target.

All pebbling formulas with a single target are minimally unsatisfiable, *i.e.* any strict subset of their clauses admits a satisfying assignment. For each formula $Pbl_G$ we use for our experiments, we also use a satisfiable version of it, called $Pbl_G^{SAT}$, obtained by randomly choosing a clause of $Pbl_G$ and deleting it. When $G$ is viewed as a task graph, $Pbl_G^{SAT}$ corresponds to a task system with a single fault, and finding a satisfying assignment for it corresponds to locating the fault.

### 2.6 The $GT_n$ Formulas

The $GT_n$ formulas are unsatisfiable CNF formulas based on the ordering principle that any partial order on the set $\{1, 2, \ldots, n\}$ must have a maximal element. They were first considered by Krishnamurthy (1985) and later used by Bonet and Galesi (2001) to show the optimality of the size-width relationship of resolution proofs. Recently, Alekhnovich et al. (2002) used a variation, called $GT'_n$, to show an exponential separation between RES and regular resolution.

The variables of $GT_n$ are $x_{i,j}$ for $i, j \in [n], i \neq j$, which should be thought of as the binary predicate $i \succ j$. Clauses $(\neg x_{i,j} \vee \neg x_{j,i})$ ensure that $\succ$ is anti-symmetric and $(\neg x_{i,j} \vee \neg x_{j,k} \vee x_{i,k})$ ensure that $\succ$ is transitive. This makes $\succ$ a partial order on $[n]$. *Successor clauses* $(\vee_{k \neq j} x_{k,j})$ provide the contradiction by saying that every element $j$ has a successor in $[n] \setminus \{j\}$, which is clearly false for the maximal elements of $[n]$ under the ordering $\succ$.

These formulas, although capturing a simple mathematical principle, are empirically difficult for many SAT solvers including zChaff. We employ our techniques to improve the performance of zChaff on these formulas. We use for our experiments the unsatisfiable version $GT_n$ described above as well as a satisfiable version $GT_n^{SAT}$ obtained by deleting a randomly chosen successor clause. The reason we consider these ordering formulas in addition to seemingly harder pebbling formulas is that the latter admit short tree-like proofs in certain extensions of RES whereas the former seem to critically require reuse of derived or learned clauses for short refutations. We elaborate on this in Section 6.2.

## 3. A Formal Framework for Studying Clause Learning

Although many SAT solvers based on clause learning have been proposed and demonstrated to be empirically successful, a theoretical discussion of the underlying concepts and structures needed for our analysis is lacking. This section focuses on this formal framework.

### 3.1 Unit Propagation and Decision Levels

All clause learning algorithms discussed in this paper are based on *unit propagation*, which is the process of repeatedly applying the unit clause rule followed by formula simplification until no clause with exactly one unassigned literal remains. In this context, it is convenient to work with residual formulas at different stages of DPLL. Let $\rho$ be the partial assignment at some stage of DPLL on formula $F$. The *residual formula* at this stage is obtained by applying unit propagation to the simplified formula $F|_\rho$.

When using unit propagation, variables assigned values through the actual branching process are called *decision* variables and those assigned values as a result of unit propagation are called *implied* variables. *Decision* and *implied literals* are analogously defined. Upon backtracking, the last decision variable no longer remains a decision variable and might instead become an implied variable depending on the clauses learned so far. The *decision level of a decision variable $x$* is one more than the number of current decision variables at the time of branching on $x$. The *decision level of an implied variable* is the maximum of the decision levels of decision variables used to imply it. The *decision level* at any step of the underlying DPLL procedure is the maximum of the decision levels of all current decision variables. Thus, for instance, if the clause learning algorithm starts off by branching on $x$, the decision level of $x$ is 1 and the algorithm at this stage is at decision level 1.

A clause learning algorithm stops and declares the given formula to be unsatisfiable whenever unit propagation leads to a conflict at decision level zero, *i.e.* when no variable is currently branched upon. This condition will be referred to in this paper as a conflict at decision level zero.

### 3.2 Branching Sequence

We use the notion of branching sequence to prove an exponential separation between DPLL and clause learning. It generalizes the idea of a static *variable order* by letting the order differ from branch to branch in the underlying DPLL procedure. In addition, it also specifies which branch (TRUE or FALSE) to explore first. This can clearly be useful for satisfiable formulas, and can also help on unsatisfiable ones by making the algorithm learn useful clauses earlier in the process.

**Definition 3.** A *branching sequence* for a CNF formula $F$ is a sequence $\sigma = (l_1, l_2, \ldots, l_k)$ of literals of $F$, possibly with repetitions. A DPLL based algorithm $\mathcal{A}$ on $F$ *branches according to $\sigma$* if it always picks the next variable $v$ to branch on in the literal order given by $\sigma$, skips $v$ if $v$ is currently assigned a value, and otherwise branches further by setting the chosen literal to FALSE and deleting it from $\sigma$. When $\sigma$ becomes empty, $\mathcal{A}$ reverts back to its default branching scheme.

**Definition 4.** A branching sequence $\sigma$ is *complete* for a formula $F$ under a DPLL based algorithm $\mathcal{A}$ if $\mathcal{A}$ branching according to $\sigma$ terminates before or as soon as $\sigma$ becomes empty. Otherwise it is *incomplete* or *approximate*.

Clearly, how well a branching sequence works for a formula depends on the specifics of the clause learning algorithm used, such as its learning scheme and backtracking process. One needs to keep these in mind when generating the sequence. It is also important to note that while the size of a variable order is always the same as the number of variables in the formula, that of an effective branching sequence is typically much more. In fact, the size of a branching sequence complete for an unsatisfiable formula $F$ is equal to the size of an unsatisfiability proof of $F$, and when $F$ is satisfiable, it is proportional to the time needed to find a satisfying assignment.

### 3.3 Clause Learning Proofs

The notion of clause learning proofs connects clause learning with resolution and provides the basis for our complexity bounds. If a given CNF formula $F$ is unsatisfiable, clause learning terminates with a conflict at decision level zero. Since all clauses used in this final conflict themselves follow directly or indirectly from $F$, this failure of clause learning in finding a satisfying assignment constitutes a logical proof of unsatisfiability of $F$. We denote by CL the proof system consisting of all such proofs. Our bounds compare the sizes of proofs in CL with the sizes of (possibly restricted) resolution proofs. Recall that clause learning algorithms can use one of many learning schemes, resulting in different proofs.

**Definition 5.** A *clause learning (CL) proof* $\pi$ of an unsatisfiable CNF formula $F$ under learning scheme $\mathcal{S}$ and induced by branching sequence $\sigma$ is the result of applying DPLL with unit propagation on $F$, branching according to $\sigma$, and using scheme $\mathcal{S}$ to learn conflict clauses such that at the end of this process, there is a conflict at decision level zero. The *size* of the proof, $size(\pi)$, is $|\sigma|$.

### 3.4 Implication Graph and Conflicts

Unit propagation can be naturally associated with an *implication graph* that captures all possible ways of deriving all implied literals from decision literals.

**Definition 6.** The *implication graph* $G$ at a given stage of DPLL is a directed acyclic graph with edges labeled with sets of clauses. It is constructed as follows:

1. Create a node for each decision literal, labeled with that literal. These will be the indegree zero source nodes of $G$.

2. While there exists a known clause $C = (l_1 \vee \ldots l_k \vee l)$ such that $\neg l_1, \ldots, \neg l_k$ label nodes in $G$,

    (a) Add a node labeled $l$ if not already present in $G$.
    
    (b) Add edges $(l_i, l), 1 \leq i \leq k$, if not already present.
    
    (c) Add $C$ to the label set of these edges. These edges are thought of as grouped together and associated with clause $C$.

3. Add to $G$ a special node $\Lambda$. For any variable $x$ which occurs both positively and negatively in $G$, add directed edges from $x$ and $\neg x$ to $\Lambda$.

Since all node labels in $G$ are distinct, we identify nodes with the literals labeling them. Any variable $x$ occurring both positively and negatively in $G$ is a *conflict variable*, and $x$ as well as $\neg x$ are *conflict literals*. $G$ contains a *conflict* if it has at least one conflict variable. DPLL at a given stage has a *conflict* if the implication graph at that stage contains a conflict. A conflict can equivalently be thought of as occurring when the residual formula contains the empty clause $\Lambda$.

By definition, an implication graph may not contain a conflict at all, or it may contain many conflict variables and several ways of deriving any single literal. To better understand and analyze a conflict when it occurs, we work with a subgraph of an implication graph, called the *conflict graph* (see Figure 2), that captures only one among possibly many ways of reaching a conflict from the decision variables using unit propagation.

**Definition 7.** A *conflict graph* $H$ is any subgraph of an implication graph with the following properties:

1. $H$ contains $\Lambda$ and exactly one conflict variable.

2. All nodes in $H$ have a path to $\Lambda$.

3. Every node $l$ in $H$ other than $\Lambda$ either corresponds to a decision literal or has precisely the nodes $\neg l_1, \neg l_2, \ldots, \neg l_k$ as predecessors where $(l_1 \vee l_2 \vee \ldots \vee l_k \vee l)$ is a known clause.

While an implication graph may or may not contain conflicts, a conflict graph always contains exactly one. The choice of the conflict graph is part of the strategy of the solver. A typical strategy will maintain one subgraph of an implication graph that has properties 2 and 3 from Definition 7, but not property 1. This can be thought of as a *unique inference* subgraph of the implication graph. When a conflict is reached, this unique inference subgraph is extended to satisfy property 1 as well, resulting in a conflict graph, which is then used to analyze the conflict.

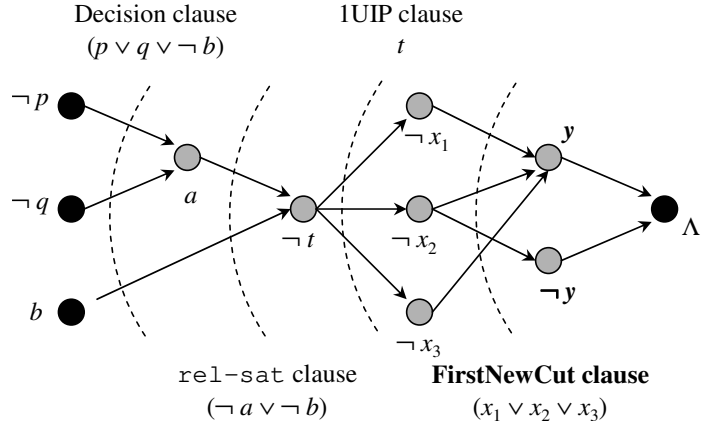

Figure 2: A conflict graph along with various learning schemes

### 3.4.1 Conflict Clauses

Consider the implication graph at a stage where there is a conflict and fix a conflict graph contained in that implication graph. Pick any cut in the conflict graph that has all decision variables on one side, called the *reason side*, and $\Lambda$ as well as at least one conflict literal on the other side, called the *conflict side*. All nodes on the reason side that have at least one edge going to the conflict side form a *cause* of the conflict. The negations of the corresponding literals forms the *conflict clause* associated with this cut.

### 3.5 Trivial Resolution and Learned Clauses

**Definition 8.** A resolution derivation $(C_1, C_2, \ldots, C_k)$ is *trivial* iff all variables resolved upon are distinct and each $C_i, i \geq 3$, is either an initial clause or is derived by resolving $C_{i-1}$ with an initial clause.

A trivial derivation is tree-like, regular, linear, as well as ordered. As the following Propositions show, trivial derivations correspond to conflicts in clause learning algorithms.

**Proposition 3.** *Let $F$ be a CNF formula. If there is a trivial resolution derivation of a clause $C \notin F$ from $F$ then setting all literals of $C$ to* FALSE *leads to a conflict by unit propagation.*

*Proof.* Let $\pi = (C_1, C_2, \ldots, C_k \equiv C)$ be a trivial resolution derivation of $C$ from $F$. Let $C_k = (l_1 \vee l_2 \vee \ldots \vee l_q)$ and $\rho$ be the partial assignment that sets all $l_i, 1 \leq i \leq q$, to FALSE. Assume without loss of generality that clauses in $\pi$ are ordered so that all initial clauses precede any derived clause. We give a proof by induction on the number of derived clauses in $\pi$.

For the base case, $\pi$ has only one derived clause, $C \equiv C_k$. Assume without loss of generality that $C_k = (A \vee B)$ and $C_k$ is derived by resolving two initial clauses $(A \vee x)$ and $(B \vee \neg x)$ on variable $x$. Since $\rho$ falsifies $C_k$, it falsifies all literals of $A$, implying $x = $ TRUE by unit propagation. Similarly, $\rho$ falsifies $B$, implying $x = $ FALSE and resulting in a conflict.

When $\pi$ has at least two derived clauses, $C_k$, by triviality of $\pi$, must be derived by resolving $C_{k-1} \notin F$ with a clause in $F$. Assume without loss of generality that $C_{k-1} \equiv$ $(A \vee x)$ and the clause from $F$ used in this resolution step is $(B \vee \neg x)$, where $C_k = (A \vee B)$. Since $\rho$ falsifies $C \equiv C_k$, it falsifies all literals of $B$, implying $x =$ FALSE by unit propagation. This in turn results in falsifying all literals of $C_{k-1}$ because all literals of $A$ are also set to FALSE by $\rho$. Now $(C_1, \ldots, C_{k-1})$ is a trivial resolution derivation of $C_{k-1} \notin F$ from $F$ with one less derived clause than $\pi$, and all literals of $C_{k-1}$ are falsified. By induction, this must lead to a conflict by unit propagation. □

**Proposition 4.** *Any conflict clause can be derived from initial and previously derived clauses using a trivial resolution derivation.*

*Proof.* Let $\sigma$ be the cut in a fixed conflict graph associated with the given conflict clause. Let $V_{conflict}(\sigma)$ denote the set of variables on the conflict side of $\sigma$, but including the conflict variable only if it occurs both positively and negatively on the conflict side. We will prove by induction on $|V_{conflict}(\sigma)|$ the stronger statement that the conflict clause associated with a cut $\sigma$ has a trivial derivation from known (*i.e.* initial or previously derived) clauses resolving precisely on the variables in $V_{conflict}(\sigma)$.

For the base case, $V_{conflict}(\sigma) = \phi$ and the conflict side contains only $\Lambda$ and a conflict literal, say $x$. The cause associated with this cut consists of node $\neg x$ that has an edge to $\Lambda$, and nodes $\neg l_1, \neg l_2, \ldots, \neg l_k$, corresponding to a known clause $C_x = (l_1 \vee l_2 \vee \ldots \vee l_k \vee x)$, that each have an edge to $x$. The conflict clause for this cut is simply the known clause $C_x$ itself, having a length zero trivial derivation.

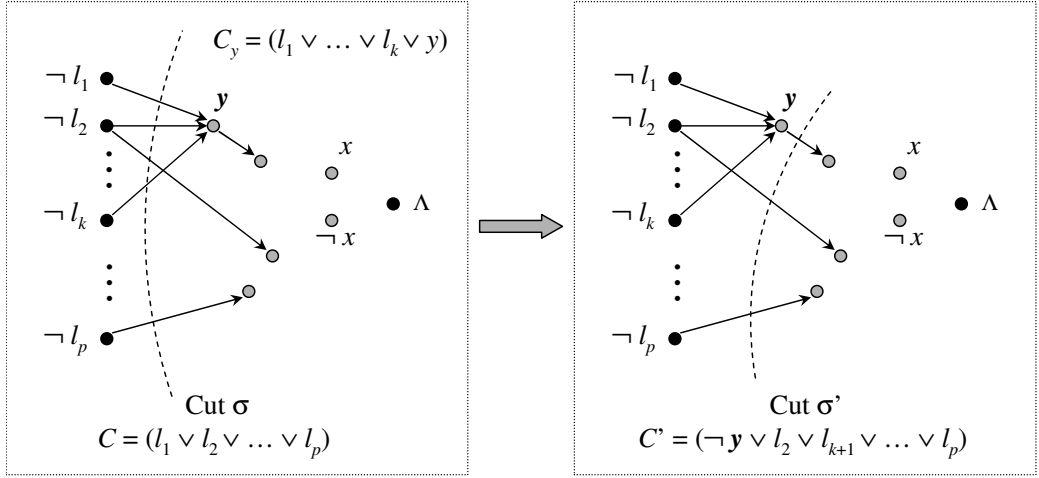

Figure 3: Deriving a conflict clause using trivial resolution. Resolving $C'$ with $C_y$ on variable $y$ gives the conflict clause $C$.

When $V_{conflict}(\sigma) \neq \phi$, pick a node $y$ on the conflict side all whose predecessors are on the reason side (see Figure. 3). Let the conflict clause be $C = (l_1 \vee l_2 \vee \ldots \vee l_p)$ and assume without loss of generality that the predecessors of $y$ are $\neg l_1, \neg l_2, \ldots, \neg l_k$ for some $k \leq p$. By definition of unit propagation, $C_y = (l_1 \vee l_2 \vee \ldots \vee l_k \vee y)$ must be a known clause. Obtain a new cut $\sigma'$ from $\sigma$ by moving node $y$ from the conflict side to the reason side. The new associated conflict clause must be of the form $C' = (\neg y \vee D)$, where $D$ is a subclause of $C$. Now $V_{conflict}(\sigma') \subset V_{conflict}(\sigma)$. Consequently, by induction, $C'$ must have a trivial resolution derivation from known clauses resolving precisely upon the variables in $V_{conflict}(\sigma')$. Recall that no variable occurs twice in a conflict graph except the conflict variable. Hence $V_{conflict}(\sigma')$ has exactly all variables of $V_{conflict}(\sigma)$ other than $y$. Using this trivial derivation of $C'$ and finally resolving $C'$ with the known clause $C_y$ on variable $y$ gives us a trivial derivation of $C$ from known clauses. This completes the inductive step. □

### 3.6 Different Learning Schemes

Different cuts separating decision variables from $\Lambda$ and a conflict literal correspond to different learning schemes (see Figure 2). One can also create learning schemes based on cuts not involving conflict literals at all (Zhang et al., 2001), but the effectiveness of such schemes is not clear. These will not be considered here.

It is insightful to think of the *nondeterministic* scheme as the most general learning scheme. Here we pick the cut nondeterministically, choosing, whenever possible, one whose associated clause is not already known. Since we can repeatedly branch on the same last variable, nondeterministic learning subsumes learning multiple clauses from a single conflict as long as the sets of nodes on the reason side of the corresponding cuts form a (set-wise) decreasing sequence. For simplicity, we will assume that only one clause is learned from any conflict.

In practice, however, we employ deterministic schemes. The *decision* scheme (Zhang et al., 2001), for example, uses the cut whose reason side comprises all decision variables. `rel-sat` (Bayardo Jr. & Schrag, 1997) uses the cut whose conflict side consists of all implied variables at the current decision level. This scheme allows the conflict clause to have exactly one variable from the current decision level, causing an automatic flip in its assignment upon backtracking.

This nice flipping property holds in general for all *unique implication points* (UIPs) (Marques-Silva & Sakallah, 1996). A UIP of an implication graph is a node at the current decision level $d$ such that any path from the decision variable at level $d$ to the conflict variable as well as its negation must go through it. Intuitively, it is a *single* reason at level $d$ that causes the conflict. Whereas `rel-sat` uses the decision variable as the obvious UIP, `GRASP` (Marques-Silva & Sakallah, 1996) and `zChaff` (Moskewicz et al., 2001) use *FirstUIP*, the one that is "closest" to the conflict variable. `GRASP` also learns multiple clauses when faced with a conflict. This makes it typically require fewer branching steps but possibly slower because of the time lost in learning and unit propagation.

The concept of UIP can be generalized to decision levels other than the current one. The *1UIP scheme* corresponds to learning the FirstUIP clause of the current decision level, the *2UIP scheme* to learning the FirstUIP clauses of both the current level and the one before, and so on. Zhang et al. (2001) present a comparison of all these and other learning schemes and conclude that 1UIP is quite robust and outperforms all other schemes they consider on most of the benchmarks.

#### 3.6.1 THE FIRSTNEWCUT SCHEME

We propose a new learning scheme called *FirstNewCut* whose ease of analysis helps us demonstrate the power of clause learning. We would like to point out that we use this scheme here only to prove our theoretical bounds using specific formulas. Its effectiveness on other formulas has not been studied yet. We would also like to point out that the experimental results in this paper are for the 1UIP learning scheme, but can also be extended to certain other schemes, including FirstNewCut.

The key idea behind FirstNewCut is to make the conflict clause as relevant to the current conflict as possible by choosing a cut close to the conflict literals. This is what the FirstUIP scheme also tries to achieve in a slightly different manner. For the following definitions, fix a cut in a conflict graph and let $S$ be the set of nodes on the reason side that have an edge to some node on the conflict side. $S$ is the reason side *frontier* of the cut. Let $C_S$ be the conflict clause associated with this cut.

**Definition 9.** *Minimization* of conflict clause $C_S$ is the following process: while there exists a node $v \in S$ all of whose predecessors are also in $S$, move $v$ to the conflict side, remove it from $S$, and repeat.

**Definition 10.** *FirstNewCut scheme*: Start with a cut whose conflict side consists of $\Lambda$ and a conflict literal. If necessary, repeat the following until the associated conflict clause, after minimization, is not already known: pick a node on the conflict side, and move all its predecessors that lie on the reason side, other than those that correspond to decision variables, to the conflict side. Finally, learn the resulting new minimized conflict clause.

This scheme starts with the cut that is closest to the conflict literals and iteratively moves it back toward the decision variables until a new associated conflict clause is found. This backward search always halts because the cut with all decision variables on the reason side is certainly a new cut. Note that there are potentially several ways of choosing a literal to move the cut back, leading to different conflict clauses. The FirstNewCut scheme, by definition, always learns a clause not already known. This motivates the following:

**Definition 11.** A clause learning scheme is *non-redundant* if on a conflict, it always learns a clause not already known.

### 3.7 Fast Backtracking and Restarts

Most clause learning algorithms use *fast backtracking* or *conflict directed backjumping* where one uses the conflict graph to undo not only the last branching decision but also all other recent decisions that did not contribute to the current conflict (Stallman & Sussman, 1977). In particular, the SAT solver `zChaff` that we use for our experiments backtracks to decision level zero when it learns a unit clause. This property influences the structure of the sequence generation algorithm presented in Section 6.1.1.

More precisely, the level that a clause learning algorithm employing this technique backtracks to is one less than the maximum of the decision levels of all decision variables (*i.e.* the *sources* of the conflict) present in the underlying conflict graph. Note that the current conflict might use clauses learned earlier as a result of branching on the apparently redundant variables. This implies that fast backtracking in general cannot be replaced by a "good" branching sequence that does not produce redundant branches. For the same reason, fast backtracking cannot either be replaced by simply learning the decision scheme clause. However, the results we present in this paper are independent of whether or not fast backtracking is used.

*Restarts* allow clause learning algorithms to arbitrarily restart their branching process from decision level zero. All clauses learned so far are however retained and now treated as additional initial clauses (Baptista & Silva, 2000). As we will show, unlimited restarts, performed at the correct step, can make clause learning very powerful. In practice, this requires extending the strategy employed by the solver to include when and how often to restart. Unless otherwise stated, clause learning proofs will be assumed to allow no restarts.

## 4. Clause Learning and Proper Natural Refinements of RES

We prove that the proof system CL, even without restarts, is stronger than *all* proper natural refinements of RES. We do this by first introducing a way of extending any CNF formula based on a given RES proof of it. We then show that if a formula $F$ $f(n)$-separates RES from a natural refinement $S$, its extension $f(n)$-separates CL from $S$. The existence of such an $F$ is guaranteed for all $f(n)$-proper natural refinements by definition.

### 4.1 The Proof Trace Extension

**Definition 12.** Let $F$ be a CNF formula and $\pi$ be a RES refutation of it. Let the last step of $\pi$ resolve $v$ with $\neg v$. Let $S = \pi \setminus (F \cup \{\neg v, \Lambda\})$. The *proof trace extension* $PT(F, \pi)$ of $F$ is a CNF formula over variables of $F$ and new trace variables $t_C$ for clauses $C \in S$. The clauses of $PT(F, \pi)$ are all initial clauses of $F$ together with a trace clause $(\neg x \vee t_C)$ for each clause $C \in S$ and each literal $x \in C$.

We first show that if a formula has a short RES refutation, then the corresponding proof trace extension has a short CL proof. Intuitively, the new trace variables allow us to simulate every resolution step of the original proof individually, without worrying about extra branches left over after learning a derived clause.

**Lemma 1.** *Suppose a formula $F$ has a RES refutation $\pi$. Let $F' = PT(F, \pi)$. Then $\mathcal{C}_{\mathsf{CL}}(F') < size(\pi)$ when CL uses the FirstNewCut scheme and no restarts.*

*Proof.* Suppose $\pi$ contains a derived clause $C_i$ whose strict subclause $C'_i$ can be derived by resolving two previously occurring clauses. We can replace $C_i$ with $C'_i$, do trivial simplifications on further derivations that used $C_i$ and obtain a simpler proof $\pi'$ of $F$. Doing this repeatedly will remove all such redundant clauses and leave us with a simplified proof no larger in size. Hence we will assume without loss of generality that $\pi$ has no such clause.

Viewing $\pi$ as a sequence of clauses, its last two elements must be a literal, say $v$, and $\Lambda$. Let $S = \pi \setminus (F \cup \{v, \Lambda\})$. Let $(C_1, C_2, \ldots, C_k)$ be the subsequence of $\pi$ that has precisely the clauses in $S$. Note that $C_i \equiv \neg v$ for some $i, 1 \leq i \leq k$. We claim that the branching sequence $\sigma = (t_{C_1}, t_{C_2}, \ldots, t_{C_k})$ induces a CL proof of $F$ of size $k$ using the FirstNewCut scheme. To prove this, we show by induction that after $i$ branching steps, the clause learning procedure branching according to $\sigma$ has learned clauses $C_1, C_2, \ldots, C_i$, has trace variables $t_{C_1}, t_{C_2}, \ldots, t_{C_i}$ set to TRUE, and is at decision level $i$.

The base case for induction, $i = 0$, is trivial. The clause learning procedure is at decision level zero and no clauses have been learned. Suppose the inductive claim holds after branching step $i - 1$. Let $C_i = (x_1 \vee x_2 \vee \ldots \vee x_l)$. $C_i$ must have been derived in $\pi$ by resolving two clauses $(A \vee y)$ and $(B \vee \neg y)$ coming from $F \cup \{C_1, C_2, \ldots, C_{i-1}\}$, where $C_i = (A \vee B)$. The $i^{th}$ branching step sets $t_{C_i} =$ FALSE. Unit propagation using trace clauses $(\neg x_j \vee t_{C_i})$, $1 \leq j \leq l$, sets each $x_j$ to FALSE, thereby falsifying all literals of $A$ and $B$. Further unit propagation using $(A \vee y)$ and $(B \vee \neg y)$ implies $y$ as well as $\neg y$, leading to a conflict. The cut in the conflict graph containing $y$ and $\neg y$ on the conflict side and everything else on the reason side yields $C_i$ as the FirstNewCut clause, which is learned from this conflict. The process now backtracks and flips the branch on $t_{C_i}$ by setting it to TRUE. At this stage, the clause learning procedure has learned clauses $C_1, C_2, \ldots, C_i$, has trace variables $t_{C_1}, t_{C_2}, \ldots, t_{C_i}$ set to TRUE, and is at decision level $i$. This completes the inductive step.

The inductive proof above shows that when the clause learning procedure has finished branching on all $k$ literals in $\sigma$, it will have learned all clauses in $S$. Adding to this the initial clauses $F$ that are already known, the procedure will have as known clauses $\neg v$ as well as the two unit or binary clauses used to derive $v$ in $\pi$. These immediately generate $\Lambda$ in the residual formula by unit propagation using variable $v$, leading to a conflict at decision level $k$. Since this conflict does not use any decision variable, fast backtracking retracts all $k$ branches. The conflict, however, still exists at decision level zero, thereby concluding the clause learning procedure and finishing the CL proof. □

**Lemma 2.** *Let $S$ be an $f(n)$-proper natural refinement of* RES *whose weakness is witnessed by a family $\{F_n\}$ of formulas. Let $\{\pi_n\}$ be the family of shortest* RES *proofs of $\{F_n\}$. Let $\{F'_n\} = \{PT(F_n, \pi_n)\}$. For* CL *using the FirstNewCut scheme and no restarts, $\mathcal{C}_S(F'_n) \geq f(n) \cdot \mathcal{C}_{\mathsf{CL}}(F'_n)$.*

*Proof.* Let $\rho_n$ the restriction that sets every trace variable of $F'_n$ to TRUE. We claim that $\mathcal{C}_S(F'_n) \geq \mathcal{C}_S(F'_n|_{\rho_n}) = \mathcal{C}_S(F_n) \geq f(n) \cdot \mathcal{C}_{\mathsf{RES}}(F_n) > f(n) \cdot \mathcal{C}_{\mathsf{CL}}(F'_n)$. The first inequality holds because $S$ is a natural proof system. The following equality holds because $\rho_n$ keeps the original clauses of $F_n$ intact and trivially satisfies all trace clauses, thereby reducing the initial clauses of $F'_n$ to precisely $F_n$. The next inequality holds because $S$ is an $f(n)$-proper refinement of RES. The final inequality follows from Lemma 1. □

This gives our first main result and its corollary using Proposition 2:

**Theorem 1.** *For any $f(n)$-proper natural refinement $S$ of* RES *and for* CL *using the FirstNewCut scheme and no restarts, there exist formulas $\{F_n\}$ such that $\mathcal{C}_S(F_n) \geq f(n) \cdot \mathcal{C}_{\mathsf{CL}}(F_n)$.*

**Corollary 1.** CL *can provide exponentially shorter proofs than tree-like, regular, and Davis-Putnam resolution.*

**Remark.** As clause learning yields resolution proofs of unsatisfiable formulas, CL is a refinement of RES. However, it is not necessarily a natural proof system. If it were shown to be natural, Theorem 1, by a contradiction argument, would imply that CL using the FirstNewCut scheme and no restarts is as powerful as RES itself.

## 5. Clause Learning and General Resolution

We begin this section by showing that CL proofs, irrespective of the learning scheme, branching strategy, or restarts used, can be efficiently simulated by RES. In the reverse direction, we show that CL, with a slight variation and with unlimited restarts, can efficiently simulate RES in its full generality. The variation relates to the variables one is allowed to branch upon.

**Lemma 3.** *For any formula $F$ over $n$ variables and CL using any learning scheme and any number of restarts, $\mathcal{C}_{\mathsf{RES}}(F) \leq n \cdot \mathcal{C}_{\mathsf{CL}}(F)$.*

*Proof.* Given a CL proof $\pi$ of $F$, a RES proof can be constructed by sequentially deriving all clauses that $\pi$ learns, which includes the empty clause $\Lambda$. From Proposition 4, all these derivations are trivial and hence require at most $n$ steps each. Consequently, the size of the resulting RES proof is at most $n \cdot size(\pi)$. Note that since we derive clauses of $\pi$ individually, restarts in $\pi$ do not affect the construction. □

**Definition 13.** Let CL-- denote the variation of CL where one is allowed to branch on a literal whose value is already set explicitly or because of unit propagation.

Of course, such a relaxation is useless in ordinary DPLL; there is no benefit in branching on a variable that doesn't even appear in the residual formula. However, with clause learning, such a branch can lead to an immediate conflict and allow one to learn a key conflict clause that would otherwise have not been learned. We will use this property to show that RES can be efficiently simulated by CL-- with enough restarts.

We first state a generalization of Lemma 3. CL-- can, by definition, do all that usual CL can, and is potentially stronger. The simulation of CL by RES can in fact be extended to CL-- as well. The proof goes exactly as the proof of Lemma 3 and uses the easy fact that Proposition 4 doesn't change even when one is allowed to branch on variables that are already set. This gives us:

**Proposition 5.** *For any formula $F$ over $n$ variables and CL-- using any learning scheme and any number of restarts, $\mathcal{C}_{\mathsf{RES}}(F) \leq n \cdot \mathcal{C}_{\mathsf{CL\text{--}}}(F)$.*

**Lemma 4.** *For any formula $F$ over $n$ variables and CL using any non-redundant scheme and at most $\mathcal{C}_{\mathsf{RES}}(F)$ restarts, $\mathcal{C}_{\mathsf{CL\text{--}}}(F) \leq n \cdot \mathcal{C}_{\mathsf{RES}}(F)$.*

*Proof.* Let $\pi$ be a RES proof of $F$ of size $s$. Assume without loss of generality as in the proof of Lemma 1 that $\pi$ does not contain a derived clause $C_i$ whose strict subclause $C'_i$ can be derived by resolving two clauses occurring previously in $\pi$. The proof of this Lemma is very similar to that of Lemma 1. However, since we do not have trace variable to allow us to simulate each resolution step individually and independently, we use explicit restarts.

Viewing $\pi$ as a sequence of clauses, its last two elements must be a literal, say $v$, and $\Lambda$. Let $S = \pi \setminus (F \cup \{v, \Lambda\})$. Let $(C_1, C_2, \ldots, C_k)$ be the subsequence of $\pi$ that has precisely the clauses in $S$. Note that $C_i \equiv \neg v$ for some $i, 1 \leq i \leq k$. For convenience, define an *extended branching sequence* to be a branching sequence in which certain places, instead of being literals, can be marked as restart points. Let $\sigma$ be the extended branching sequence consisting of all literals of $C_1$, followed by a restart point, followed by all literals of $C_2$, followed by a second restart point, and so on up to $C_k$. We claim that $\sigma$ induces a CL-- proof of $F$ using any non-redundant learning scheme. To prove this, we show by induction that after the $i^{th}$ restart point in $\sigma$, the CL-- procedure has learned clauses $C_1, C_2, \ldots, C_i$ and is at decision level zero.

The base case for induction, $i = 0$, is trivial. No clauses have been learned and the clause learning procedure is at decision level zero. Suppose the inductive claim holds after the $(i-1)^{st}$ restart point in $\sigma$. Let $C_i = (x_1 \vee x_2 \vee \ldots \vee x_l)$. $C_i$ must have been derived in $\pi$ by resolving two clauses $(A \vee y)$ and $(B \vee \neg y)$ coming from $F \cup \{C_1, C_2, \ldots, C_{i-1}\}$, where $C_i = (A \vee B)$. Continuing to branch according to $\sigma$ till before the $i^{th}$ restart point makes the CL-- procedure set all if $x_1, x_2, \ldots, x_l$ to FALSE. Note that when all literals appearing in $A$ and $B$ are distinct, the last branch on $x_l$ here is on a variable that is already set because of unit propagation. CL--, however, allows this. At this stage, unit propagation using $(A \vee y)$ and $(B \vee \neg y)$ implies $y$ as well as $\neg y$, leading to a conflict. The conflict graph consists of $\neg x_j$'s, $1 \leq j \leq l$, as the decision literals, $y$ and $\neg y$ as implied literals, and $\Lambda$. The only new conflict clause that can learned from this very simple conflict graph is $C_i$. Thus, $C_i$ is learned using any non-redundant learning scheme and the $i^{th}$ restart executed, as dictated by $\sigma$. At this stage, the CL-- procedure has learned clauses $C_1, C_2, \ldots, C_i$, and is at decision level zero. This completes the inductive step.

The inductive proof above shows that when the CL-- procedure has finished with the $k^{th}$ restart in $\sigma$, it will have learned all clauses in $S$. Adding to this the initial clauses $F$ that are already known, the procedure will have as known clauses $\neg v$ as well as the two unit or binary clauses used to derive $v$ in $\pi$. These immediately generate $\Lambda$ in the residual formula by unit propagation using variable $v$, leading to a conflict at decision level zero, thereby concluding the clause learning procedure and finishing the CL-- proof. The bounds on the size of this proof and the number of restarts needed immediately follow from the definition of $\sigma$. □

Combining Lemma 4 with Proposition 5, we get

**Theorem 2.** CL-- *with any non-redundant scheme and unlimited restarts is polynomially equivalent to* RES.

Note that Baptista and Silva (2000) showed that CL together with restarts is *complete*. Our theorem makes a much stronger claim about a slight variation of CL, namely, with enough restarts, this variation can always find proofs that are as short as those of RES.

## 6. From Analysis to Practice

The complexity bounds established in the previous sections indicate that clause learning is potentially quite powerful, especially when compared to ordinary DPLL. However, natural choices such as which conflict graph to choose, which cut in it to consider, in what order to branch on variables, and when to restart, make the process highly nondeterministic. These choices must be made deterministically (or randomly) when implementing a clause learning algorithm. To harness its full potential on a given problem domain, one must, in particular, implement a learning scheme and a branch decision process suited to that domain.

### 6.1 Solving Pebbling Formulas

As a first step toward our grand goal of translating theoretical understanding into effective implementations, we show, using pebbling problems as a concrete example, how one can utilize high level problem descriptions to generate effective branching strategies for clause

BEAME, KAUTZ, & SABHARWAL

learning algorithms. Specifically, we use insights from our theoretical analysis to give an efficient algorithm to generate an effective branching sequence for unsatisfiable as well as satisfiable pebbling formulas (see Section 2.5). This algorithm takes as input the underlying pebbling graph (which is the high level description of the pebbling problem), and not the CNF pebbling formula itself. As we will see in Section 6.3, the generated branching sequence gives exponential empirical speedup over zChaff for both grid and randomized pebbling formulas.

zChaff, despite being one of the current best clause learners, by default does not perform very well on seemingly simple pebbling formulas, even on the uniform grid version. Although clause learning should ideally need only polynomial time to solve these problem instances (in fact, linear time in the size of the formula), choosing a good branching order is critical for this to happen. Since nodes are intuitively pebbled in a bottom up fashion, we must also learn the right clauses (*i.e.* clauses labeling the nodes) in a bottom up order. However, branching on variables labeling lower nodes before those labeling higher ones prevents any DPLL based learning algorithm from backtracking the right distance and proceeding further in an effective manner. To make this clear, consider the general pebbling graph of Figure 1. Suppose we branch on and set $d_1, d_2, d_3$ and $a_1$ to FALSE. This will lead to a contradiction through unit propagation by implying $a_2$ is TRUE and $b_1$ and $b_2$ are both FALSE. We will learn $(d_1 \vee d_2 \vee d_3 \vee \neg a_2)$ as the associated 1UIP conflict clause and backtrack. There will still be a contradiction without any further branches, making us learn $(d_1 \vee d_2 \vee d_3)$ and backtrack. At this stage, we will have learned the correct clause but will be *stuck* with two branches on $d_1$ and $d_2$. Unless we had branched on $e_1$ before branching on the variables of node $d$, we will not be able to learn $e_1$ as the clause corresponding to the next higher pebbling node.

### 6.1.1 AUTOMATIC SEQUENCE GENERATION: PebSeq1UIP

Algorithm 1, PebSeq1UIP, describes a way of generating a good branching sequence for pebbling formulas. It works on any pebbling graph $G$ with distinct label variables as input and produces a branching sequence linear in the size of the associated pebbling formula. In particular, the sequence size is linear in the number of variables as well when the indegree as well as label size are bounded by a constant.

PebSeq1UIP starts off by handling the set $U$ of all nodes labeled with unit clauses. Their outgoing edges are deleted and they are treated as pseudo sources. The procedure first generates a branching sequence for non-target nodes in $U$ in increasing order of height. The key here is that when zChaff learns a unit clause, it fast-backtracks to decision level zero, effectively restarting at that point. We make use of this fact to learn these unit clauses in a bottom up fashion, unlike the rest of the process which proceeds top down in a depth-first way.

PebSeq1UIP now adds branching sequences for the targets. Note that for an unsatisfiability proof, we only need the sequence corresponding to the first (lowest) target. However, we process all targets so that this same sequence can also be used when the formula is made satisfiable by deleting enough clauses. The subroutine PebSubseq1UIP runs on a node $v$, looking at its $i^{th}$ predecessor $u$ in increasing order by height. No labels are output if $u$ is the lowest predecessor; the negations of these variables will be indirectly implied during

**Input** : Pebbling graph $G$ with no repeated labels
**Output** : Branching sequence for $Pbl_G$ for the 1UIP learning scheme
**begin**

    **foreach** $v$ *in* `BottomUpTraversal(`$G$`)` **do**
        $v.height \leftarrow 1 + max_{u \in v.preds}\{u.height\}$
        `Sort(`$v.preds$, *increasing order by height*`)`

    // First handle unit clause labeled nodes and generate their sequence
    $U \leftarrow \{v \in G.nodes : |v.labels| = 1\}$
    $G.edges \leftarrow G.edges \setminus \{(u,v) \in G.edges : u \in U\}$
    Add to $G.sources$ any new nodes with now 0 preds
    `Sort(`$U$, *increasing order by height*`)`
    **foreach** $u \in U \setminus G.targets$ **do**
        Output $u.label$
        `PebSubseq1UIPWrapper(`$u$`)`

    // Now add branching sequence for targets by increasing height
    `Sort(`$G.targets$, *increasing order by height*`)`
    **foreach** $t \in G.targets$ **do**
        `PebSubseq1UIPWrapper(`$t$`)`

**end**

`PebSubseq1UIPWrapper(`*node* $v$`)` **begin**
    **if** $|v.preds| > 0$ **then**
        `PebSubseq1UIP(`$v$, $|v.preds|$`)`
**end**

`PebSubseq1UIP(`*node* $v$, *int* $i$`)` **begin**
    $u \leftarrow v.preds[i]$

    // If this is the lowest predecessor ...
    **if** $i = 1$ **then**
        **if** $!u.visited$ **and** $u \notin G.sources$ **then**
            $u.visited \leftarrow$ TRUE
            `PebSubseq1UIPWrapper(`$u$`)`
        **return**

    // If this is not the lowest one ...
    Output $u.labels \setminus \{u.lastLabel\}$
    **if** $!u.visitedAsHigh$ **and** $u \notin G.sources$ **then**
        $u.visitedAsHigh \leftarrow$ TRUE
        Output $u.lastLabel$
        **if** $!u.visited$ **then**
            $u.visited \leftarrow$ TRUE
            `PebSubseq1UIPWrapper(`$u$`)`

    `PebSubseq1UIP(`$v, i-1$`)`

    **for** $j \leftarrow (|u.labels| - 2)$ ***downto*** $1$ **do**
        Output $u.labels[1], \ldots, u.labels[j]$
        `PebSubseq1UIP(`$v, i-1$`)`
    `PebSubseq1UIP(`$v, i-1$`)`
**end**

Algorithm 1: `PebSeq1UIP`, generating branching sequence for pebbling formulas

clause learning. However, it is recursed upon if not previously visited. This recursive sequence results in learning something close to the clause labeling this lowest node, but not quite that exact clause. If $u$ is a higher predecessor (it will be marked as *visitedAsHigh*), `PebSubseq1UIP` outputs all but one variables labeling $u$. If $u$ is not a source and has not previously been visited as high, the last label is output as well, and $u$ recursed upon if necessary. This recursive sequence results in learning the clause labeling $u$. Finally, `PebSubseq1UIP` generates a recursive pattern, calling the subroutine with the next lower predecessor of $v$. The precise structure of this pattern is dictated by the 1UIP learning scheme and fast backtracking used in `zChaff`. Its size is exponential in the degree of $v$ with label size as the base.

**The Grid Case.** It is insightful to look at the simplified version of the sequence generation algorithm that works only for grid pebbling formulas. This is described below as Algorithm 2, `GridPebSeq1UIP`. Note that both predecessors of any node are at the same level for grid pebbling graphs and need not be sorted by height. There are no nodes labeled with unit clauses and there is exactly one target node $t$, simplifying the whole algorithm to a single call to `PebSubseq1UIP(t,2)` in the notation of Algorithm 1. The last *for* loop of this procedure and the recursive call that follows it are now redundant. We combine the original wrapper method and the calls to `PebSubseq1UIP` with parameters $(v,2)$ and $(v,1)$ into a single method `GridPebSubseq1UIP` with parameter $v$.

```
Input   : Grid pebbling graph G with target node t
Output  : Branching sequence for Pbl_G for the 1UIP learning scheme
begin
    GridPebSubseq1UIP(t)
end

GridPebSubseq1UIP(node v) begin
    if v ∈ G.sources then
        return

    u ← v.preds.left
    Output u.firstLabel
    if !u.visitedAsLeft and u ∉ G.sources then
        u.visitedAsLeft ← TRUE
        Output u.secondLabel
        if !u.visited then
            u.visited ← TRUE
            GridPebSubseq1UIP(u)

    u ← v.preds.right
    if !u.visited and u ∉ G.sources then
        u.visited ← TRUE
        GridPebSubseq1UIP(u)
end
```

Algorithm 2: `GridPebSeq1UIP`, generating branching sequence for grid pebbling formulas

The resulting branching sequence can actually be generated by a simple depth first traversal (DFS) of the grid pebbling graph, printing no labels for the nodes on the rightmost path (including the target node), both labels for internal nodes, and one arbitrarily chosen label for source nodes. However, this resemblance to DFS is a somewhat misleading coincidence. The resulting sequence diverges substantially from DFS order as soon as label size or indegree of some nodes is changed. For the 10 node depth 4 grid pebbling graph shown in Figure 1, the branching sequence generated by the algorithm is $h_1, h_2, e_1, e_2, a_1, b_1, f_1, f_2, c_1$. Here, for instance, $b_1$ is generated after $a_1$ not because it labels the right (second) predecessor of node $e$ but because it labels the left (first) predecessor of node $f$. Similarly, $f_1$ and $f_2$ appear after the subtree rooted at $h$ as left predecessors of node $i$ rather than as right predecessors of node $h$.

**Example for the General Case**. To clarify the general algorithm, we describe its execution on a small example. Let $G$ be the pebbling graph in Figure 4. Denote by $t$ the node labeled $(t_1 \vee t_2)$, and likewise for other nodes. Nodes $c, d, f$ and $g$ are at height 1, nodes $a$ and $e$ at height 2, node $b$ at height 3, and node $t$ at height 4. $U = \{a, b\}$. The edges $(a, t)$ and $(b, t)$ originating from these unit clause labeled nodes are removed, and $t$, with no predecessors anymore, is added to the list of sources. We output the label of the non-target unit nodes in $U$ in increasing order of height, and recurse on each of them in order, i.e. we output $a_1$, setting $B = (a_1)$, call `PebSubseq1UIPWrapper` on $a$, and then repeat this process for $b$. This is followed by a recursive call to `PebSubseq1UIPWrapper` on the target node $t$.

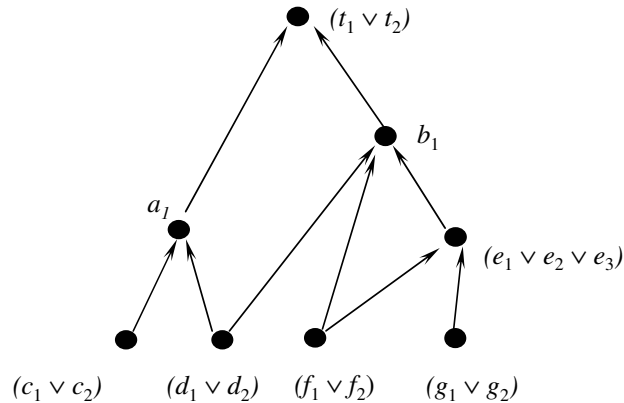

Figure 4: A simple pebbling graph to illustrate branch sequence generation

The call `PebSubseq1UIPWrapper` on $a$ in turn invokes `PebSubseq1UIP` with parameters $(a, 2)$. This sorts the predecessors of $a$ in increasing order of height to, say, $d, c$, with $d$ being the lowest predecessor. $v$ is set to $a$ and $u$ is set to the second predecessor $c$. We output all but the last label of $u$, i.e. of $c$, making the current branching sequence $B = (a_1, c_1)$. Since $u$ is a source, nothing more needs to be done for it and we make a recursive call to `PebSubseq1UIP` with parameters $(a, 1)$. This sets $u$ to $d$, which is the lowest predecessor and requires nothing to be done because it is also a source. This finishes the sequence generation for $a$, ending at $B = (a_1, c_1)$. After processing this part of the sequence, `zChaff` will have $a$ as a learned clause.

We now output $b_1$, the label of the unit clause $b$. The call, `PebSubseq1UIPWrapper` on $b$, proceeds similarly, setting predecessor order as $(d, f, e)$, with $d$ as the lowest predecessor. Procedure `PebSubseq1UIP` is called first with parameters $(b, 3)$, setting $u$ to $e$. This adds all but the last label of $e$ to the branching sequence, making it $B = (a_1, c_1, b_1, e_1, e_2)$.

Since this is the first time $e$ is being visited as high, its last label is also added, making $B = (a_1, c_1, b_1, e_1, e_2, e_3)$, and it is recursed upon with PebSubseq1UIPWrapper(e). This recursion extends the sequence to $B = (a_1, c_1, b_1, e_1, e_2, e_3, f_1)$. After processing this part of $B$, zChaff will have both $a$ and $(e_1 \vee e_2 \vee e_3)$ as learned clauses. Getting to the second highest predecessor $f$ of $b$, which happens to be a source, we simply add another $f_1$ to $B$. Finally, we get to the third highest predecessor $d$ of $b$, which happens to be the lowest as well as a source, thus requiring nothing to be done. Coming out of the recursion, back to $u = f$, we generate the pattern given by the last for loop, which is empty because the label size of $f$ is only 2. Coming out once more of the recursion to $u = e$, the for loop pattern generates $e_1, f_1$ and is followed by a call to PebSubseq1UIP with the next lower predecessor $f$ as the second parameter, which generates $f_1$. This makes the current sequence $B = (a_1, c_1, b_1, e_1, e_2, e_3, f_1, f_1, e_1, f_1, f_1)$. After processing this, zChaff will also have $b$ as a learned clause.

The final call to PebSubseq1UIPWrapper with parameter $t$ doesn't do anything because both predecessors of $t$ were removed in the beginning. Since both $a$ and $b$ have been learned, zChaff will have an immediate contradiction at decision level zero. This gives us the complete branching sequence $B = (a_1, c_1, b_1, e_1, e_2, e_3, f_1, f_1, e_1, f_1, f_1)$ for the pebbling formula $Pbl_G$.

### 6.1.2 Complexity of Sequence Generation

Let graph $G$ have $n$ nodes, indegree of non-source nodes between $d_{min}$ and $d_{max}$, and label size between $l_{min}$ and $l_{max}$. For simplicity of analysis, we will assume that $l_{min} = l_{max} = l$ and $d_{min} = d_{max} = d$ ($l = d = 2$ for a grid graph).

Let us first compute the size of the pebbling formula associated with $G$. The running time of PebSeq1UIP and the size of the branching sequence generated will be given in terms of this size. The number of clauses in the pebbling formula $Pbl_G$ is roughly $nl^d$. Taking clause sizes into account, the size of the formula, $|Pbl_G|$, is roughly $n(l+d)l^d$. Note that the size of the CNF formula itself grows exponentially with the indegree and gets worse as label size increases. The best case is when $G$ is the grid graph, where $|Pbl_G| = \Theta(n)$. This explains the degradation in performance of zChaff, both original and modified, as we move from grid graphs to random graphs (see section 6.3). Since we construct $Pbl_G^{SAT}$ by deleting exactly one randomly chosen clause from $Pbl_G$ (see Section 2.5), the size $|Pbl_G^{SAT}|$ of the satisfiable version is also essentially the same.

Let us now compute the running time of PebSeq1UIP. Initial computation of heights and predecessor sorting takes time $\Theta(nd \log d)$. Assuming $n_u$ unit clause labeled nodes and $n_t$ target nodes, the remaining node sorting time is $\Theta(n_u \log n_u + n_t \log n_t)$. Since PebSubseq1UIPWrapper is called at most once for each node, the total running time of PebSeq1UIP is $\Theta(nd \log d + n_u \log n_u + n_t \log n_t + nT_{wrapper})$, where $T_{wrapper}$ denotes the running time of PebSubseq1UIP- Wrapper without taking into account recursive calls to itself. When $n_u$ and $n_t$ are much smaller than $n$, which we will assume as the typical case, this simplifies to $\Theta(nd \log d + nT_{wrapper})$. If $T(v, i)$ denotes the running time of PebSubseq1UIP(v,i), again without including recursive calls to the wrapper method, then $T_{wrapper} = T(v, d)$. However, $T(v, d) = lT(v, d-1) + \Theta(l)$, which gives $T_{wrapper} = T(v, d) =$ $\Theta(l^{d+1})$. Substituting this back, we get that the running time of `PebSeq1UIP` is $\Theta(nl^{d+1})$, which is about the same as $|Pbl_G|$.

Finally, we consider the size of the branching sequence generated. Note that for each node, most of its contribution to the sequence is from the recursive pattern generated near the end of `PebSubseq1UIP`. Let $Q(v, i)$ denote this contribution. $Q(v, i) = (l-2)(Q(v, i-1) + \Theta(l))$, which gives $Q(v, i) = \Theta(l^{d+2})$. Hence, the size of the sequence generated is $\Theta(nl^{d+2})$, which again is about the same as $|Pbl_G|$.

**Theorem 3.** *Given a pebbling graph $G$ with label size at most $l$ and indegree of non-source nodes at most $d$, algorithm `PebSeq1UIP` produces a branching sequence $\sigma$ of size at most $S$ in time $\Theta(dS)$, where $S = |Pbl_G| \approx |Pbl_G^{SAT}|$. Moreover, the sequence $\sigma$ is complete for $Pbl_G$ as well as for $Pbl_G^{SAT}$ under any clause learning algorithm using fast backtracking and 1UIP learning scheme (such as `zChaff`).*

*Proof.* The size and running time bounds follow from the previous discussion in this section. That this sequence is complete can be verified by a simple hand calculation simulating clause learning with fast backtracking and 1UIP learning scheme. □

### 6.2 Solving $GT_n$ Formulas

While pebbling formulas are not so easy to solve by popular SAT solvers, they are inherently not too difficult for clause learning algorithms. In fact, even without any learning, they admit tree-like proofs under a somewhat stronger related proof system, RES($k$), for large enough $k$:

**Proposition 6 (Esteban et al., 2002).** *$Pbl_G$ has a size $O(|G|)$ tree-like RES($k$) refutation, where $k$ is the maximum width of a clause labeling a node of $G$. In particular, when $G$ is a grid graph with $n$ nodes, $Pbl_G$ has an $O(n)$ size tree-like RES(2) refutation.*

Here RES($k$) denotes the extension of RES that allows resolving, instead of clauses, disjunctions of conjunctions of up to $k$ literals (Krajíček, 2001). RES(1) is simply RES. Proposition 6 implies that addition of natural extension variables corresponding to $k$-conjunctions of variables of $Pbl_G$ leads to $O(|G| \cdot k)$ size tree-like resolution proofs of related pebbling formulas $Pbl_G(k)$ (Atserias & Bonet, 2002).

For $GT_n$ formulas, however, no such short tree-like proofs are known in RES($k$) for any $k$. Reusing derived clauses (equivalently, learning clauses with DPLL) seems to be the key to finding short proofs of $GT_n$. This makes them a good candidate for testing clause learning based SAT solvers. Our experiments indicate that $GT_n$ formulas, despite their simplicity, are quite hard for default `zChaff`. Using a good branching sequence based on the ordering structure underlying these formulas leads to significant performance gains. Recall that `PebSeq1UIP` was a fairly complex algorithm that generated a perfect branching sequence for randomized pebbling graphs. In contrast, the branching sequence we give below for $GT_n$ formulas is generated by a nearly trivial algorithm. It is an incomplete sequence (see Definition 4) but boosts performance in practice.

#### 6.2.1 Automatic Sequence Generation: `GTnSeq1UIP`

Since there is exactly one, well defined, unsatisfiable $GT$ formula for a fixed parameter $n$, the approximate branching sequence given in Figure 5 below is straightforward. However, the fact that the same branching sequence works well for the satisfiable version of the $GT_n$ formulas, obtained by deleting a randomly chosen successor clause, is worth noting.

$$\begin{array}{cccccc}
- & x_{2,1} & x_{3,1} & x_{4,1} & \ldots & x_{n-1,1} \\
x_{1,2} & - & x_{3,2} & x_{4,2} & \ldots & x_{n-1,2} \\
x_{1,3} & x_{2,3} & - & x_{4,3} & \ldots & x_{n-1,3} \\
x_{1,4} & x_{2,4} & x_{3,4} & - & \ldots & x_{n-1,4} \\
\vdots & & & & & \\
x_{1,n} & x_{2,n} & x_{3,n} & x_{4,n} & \ldots & - \\
x_{1,n} & x_{2,n} & x_{3,n} & x_{4,n} & \ldots & x_{n-1,n}
\end{array}$$

Figure 5: Approximate branching sequence for $GTn$ formulas. The sequence goes left to right along row 1, then along row 2, and so on. Entries marked '−' correspond to non-existent variables $x_{i,i}$.

### 6.3 Experimental Results

We conducted experiments on a Linux machine with a 1600 MHz AMD Athelon processor, 256 KB cache and 1024 MB RAM. Time limit was set to 6 hours and memory limit to 512 MB; the program was set to abort as soon as either of these was exceeded. We took the base code of zChaff (Moskewicz et al., 2001), version 2001.6.15, and modified it to incorporate a branching sequence given as part of the input, along with a CNF formula. When an incomplete branching sequence is specified that gets exhausted before a satisfying assignment is found or the formula is proved to be unsatisfiable, the code reverts to the default variable selection scheme VSIDS of zChaff. For consistency, we analyzed the performance with random restarts turned off. For all other parameters, we used the default values of zChaff. For all formulas, results are reported for DPLL (zChaff with clause learning disabled), for CL (unmodified zChaff), and for CL with a specified branching sequence (modified zChaff).

Tables 1 and 2 show the performance on grid pebbling and randomized pebbling formulas, respectively. In both cases, the branching sequence used was generated by Algorithm 1, PebSeq1UIP, and the size of problems that can be solved increases substantially as we move down the tables. Note that randomized pebbling graphs typically have a more complex structure than grid pebbling graphs. In addition, higher indegree and larger disjunction labels make both the CNF formula size as well as the required branching sequence larger. This explains the difference between the performance of zChaff, both original and modified, on grid and randomized pebbling instances. For all instances considered, the time taken to generate the branching sequence from the input graph was significantly less than that for generating the pebbling formula itself.

Table 3 shows the performance on the $GT_n$ formulas using the branching sequence given in Figure 5. As this sequence is incomplete, the solver had to revert back to zChaff's VSIDS heuristic to choose variables to branch on after using the given branching sequence as a guide for the first few decisions. Nevertheless, the sizes of problems that could be handled increased significantly. The satisfiable versions proved to be relatively easier, with or without a specified branching sequence.

Table 1: `zChaff` on *grid pebbling* formulas. ‡ denotes out of memory.

| Solver | Grid formula | | Runtime in seconds | |
|---|---|---|---|---|
| | Layers | Variables | Unsatisfiable | Satisfiable |
| DPLL | 5 | 30 | 0.24 | 0.12 |
| | 6 | 42 | 110 | 0.02 |
| | 7 | 56 | > 6 hrs | 0.07 |
| | 8 | 72 | > 6 hrs | > 6 hrs |
| CL (unmodified zChaff) | 20 | 420 | 0.12 | 0.05 |
| | 40 | 1,640 | 59 | 36 |
| | 65 | 4,290 | ‡ | 47 |
| | 70 | 4,970 | ‡ | ‡ |
| CL + **branching sequence** | 100 | 10,100 | 0.59 | 0.62 |
| | 500 | 250,500 | 254 | 288 |
| | 1,000 | 1,001,000 | 4,251 | 5,335 |
| | 1,500 | 2,551,500 | 21,097 | ‡ |

Table 2: `zChaff` on *randomized pebbling* formulas with distinct labels, indegree $\leq 5$, and disjunction label size $\leq 6$. ‡ denotes out of memory.

| Solver | Randomized pebbling formula | | | Runtime in seconds | |
|---|---|---|---|---|---|
| | Nodes | Variables | Clauses | Unsatisfiable | Satisfiable |
| DPLL | 9 | 33 | 300 | 0.00 | 0.00 |
| | 10 | 29 | 228 | 0.58 | 0.00 |
| | 10 | 48 | 604 | > 6 hrs | > 6 hrs |
| CL (unmodified zChaff) | 50 | 154 | 3,266 | 0.91 | 0.03 |
| | 87 | 296 | 9,850 | ‡ | 65 |
| | 109 | 354 | 11,106 | 584 | 0.78 |
| | 110 | 354 | 18,467 | ‡ | ‡ |
| CL + **branching sequence** | 110 | 354 | 18,467 | 0.28 | 0.29 |
| | 4,427 | 14,374 | 530,224 | 48 | 49 |
| | 7,792 | 25,105 | 944,846 | 181 | > 6 hrs |
| | 13,324 | 43,254 | 1,730,952 | 669 | 249 |

Table 3: zChaff on $GT_n$ formulas. ‡ denotes out of memory.

| Solver | $GT_n$ formula | | | Runtime in seconds | |
|---|---|---|---|---|---|
| | $n$ | Variables | Clauses | Unsatisfiable | Satisfiable |
| DPLL | 8 | 62 | 372 | 1.05 | 0.34 |
| | 9 | 79 | 549 | 48.2 | 0.82 |
| | 10 | 98 | 775 | 3395 | 248 |
| | 11 | 119 | 1,056 | > 6 hrs | 743 |
| CL (unmodified zChaff) | 10 | 98 | 775 | 0.20 | 0.00 |
| | 13 | 167 | 1,807 | 93.7 | 7.14 |
| | 15 | 223 | 2,850 | 1492 | 0.01 |
| | 18 | 322 | 5,067 | ‡ | ‡ |
| CL + branching sequence | 18 | 322 | 5,067 | 0.52 | 0.13 |
| | 27 | 727 | 17,928 | 701 | 0.17 |
| | 35 | 1,223 | 39,900 | 3.6 | 0.15 |
| | 45 | 2,023 | 86,175 | ‡ | 0.81 |

## 7. Conclusion

This paper has begun the task of formally analyzing the power of clause learning from a proof complexity perspective. Understanding where clause learning stands in relation to well studied proof systems should lead to better insights on why it works well on certain domains and fails on others. For instance, pebbling problems are an example of a domain where our results say that learning is necessary and sufficient, given a good branching order, to obtain sub-exponential solutions using DPLL based methods. On the other hand, the connection with resolution also implies that any problem that contains as a sub-problem a formula that is inherently hard even for RES, such as the pigeonhole principle (Haken, 1985), must be hard for any variant of clause learning. For such domains, theoretical results suggest practical extensions such as symmetry breaking and counting techniques for obtaining efficient solutions.

The complexity results in this paper are about proofs of unsatisfiability, and hence apply directly only to the unsatisfiable version of the pebbling formulas. Despite this, the experiments show a clear speed-up on satisfiable versions as well. This, as mentioned in Section 1, can be explained by the idea of Achlioptas et al. (2001): any DPLL based algorithm run on a satisfiable problem instance will take a long time to run precisely when the algorithm encounters an unsatisfiable sub-problem of the original problem on which it must take a long time. This lets one translate hardness results from unsatisfiable formulas to their satisfiable counterparts.

This paper inspires but leaves open several interesting questions of proof complexity. We showed that there are formulas on which CL is much more efficient than any proper natural refinement of RES. In general, can every short refutation in any such refinement be converted into a short CL proof? Or are these refinements and CL incomparable? We have shown that with arbitrary restarts, a slight variant of CL is as powerful as RES. However, judging when to restart and deciding what branching sequence to use after restarting adds more nondeterminism to the process, making it harder for practical implementations. Can CL with limited restarts also simulate RES efficiently?

In practice, a solver must employ good branching heuristics as well as implement a powerful proof system. Our results that pebbling formulas have short CL proofs depends critically upon the solver choosing a branching sequence that solves the formula in a "bottom-up" fashion, so that the learned clauses have maximal reuse. Nevertheless, we were able to automatically generate such sequences for grid and randomized pebbling formulas. Although somewhat artificial and capturing the narrow domain of task precedence, pebbling graphs are structurally similar to the layered graphs induced naturally by problems involving unwinding of state space over time, such as STRIPS planning (Kautz & Selman, 1992) and bounded model checking (Biere et al., 1999b). Pebbling problems also provide hard instances for some of the best existing SAT solvers like zChaff. This bolsters our belief that high level structure can be recovered and exploited to make clause learning more efficient.

The form in which we extract and use problem structure is a branching sequence. Although capable of capturing more information than a static variable order and avoiding the overhead of dynamic branching schemes, the exactness and detail branching sequences seem to require for pebbling formulas might pose problems when we move to harder domains where a polynomial size sequence is unlikely to exist. We may still be able to obtain substantial (but not exponential) improvements as long as an incomplete or approximate branching sequence made correct decisions most of the time, especially near the top of the underlying DPLL tree. The performance gains reported for $GT_n$ formulas indicate that even a very simple and partial branching sequence can make a big difference in practice. Along these lines, variable orders in general have been studied in other scenarios, such as for algorithms based on BDDs (see *e.g.,* Aziz et al., 1994; Harlow & Brglez, 1998). There has been work on using BDD variable orders for DPLL algorithms without learning (Reda et al., 2002). The ideas here can potentially provide new ways of capturing structural information.

Finally, our approach of exploiting high level problem description to generate auxiliary information for SAT solvers requires the knowledge of this high level description. The standard CNF benchmarks, unfortunately, do not come with such a description. Of course, there is no reason for this information to not be available since CNF formulas for practically all real-world problems are created from a more abstract specification.

## Acknowledgments

The authors wish to thank the anonymous referees for providing useful comments and for pointing out the existence of short tree-like RES($k$) proofs of pebbling formulas. This research was supported by NSF Grant ITR-0219468 and parts of this paper appeared earlier in IJCAI '03 and SAT '03 conferences (Beame et al., 2003b; Sabharwal et al., 2003).

Towards Understanding and Harnessing the Potential of Clause LearningBonet, M. L., & Galesi, N. (2001). Optimality of size-width tradeoffs for resolution. *Computational Complexity*, *10*(4), 261–276.

Brafman, R. I. (2001). A simplifier for propositional formulas with many binary clauses. In *Proceedings of the 17th International Joint Conference on Artificial Intelligence*, pp. 515–522, Seattle, WA.

Buresh-Oppenheim, J., & Pitassi, T. (2003). The complexity of resolution refinements. In *18th Annual IEEE Symposium on Logic in Computer Science*, pp. 138–147, Ottawa, Canada.

Cook, S. A., & Reckhow, R. A. (1977). The relative efficiency of propositional proof systems. *Journal of Symbolic Logic*, *44*(1), 36–50.

Davis, M., Logemann, G., & Loveland, D. (1962). A machine program for theorem proving. *Communications of the ACM*, *5*, 394–397.

Davis, M., & Putnam, H. (1960). A computing procedure for quantification theory. *Communications of the ACM*, *7*, 201–215.

Davis, R. (1984). Diagnostic reasoning based on structure and behavior. *Artificial Intelligence*, *24*, 347–410.

de Kleer, J., & Williams, B. C. (1987). Diagnosing multiple faults. *Artificial Intelligence*, *32*(1), 97–130.

Esteban, J. L., Galesi, N., & Messner, J. (2002). On the complexity of resolution with bounded conjunctions. In *Automata, Languages, and Programming: 29th International Colloquium*, Vol. 2380 of *Lecture Notes in Computer Science*, pp. 220–231, Malaga, Spain. Springer-Verlag.

Genesereth, R. (1984). The use of design descriptions in automated diagnosis. *Artificial Intelligence*, *24*, 411–436.

Giunchiglia, E., Maratea, M., & Tacchella, A. (2002). Dependent and independent variables in propositional satisfiability. In *Proceedings of the 8th European Conference on Logics in Artificial Intelligence (JELIA)*, Vol. 2424 of *Lecture Notes in Computer Science*, pp. 296–307, Cosenza, Italy. Springer-Verlag.

Gomes, C. P., Selman, B., & Kautz, H. (1998a). Boosting combinatorial search through randomization. In *Proceedings, AAAI-98: 15th National Conference on Artificial Intelligence*, pp. 431–437, Madison, WI.

Gomes, C. P., Selman, B., McAloon, K., & Tretkoff, C. (1998b). Randomization in backtrack search: Exploiting heavy-tailed profiles for solving hard scheduling problems. In *Proceedings of the 4th International Conference on Artificial Intelligence Planning Systems*, Pittsburgh, PA.

Haken, A. (1985). The intractability of resolution. *Theoretical Computer Science*, *39*, 297–305.

Harlow, J. E., & Brglez, F. (1998). Design of experiments in BDD variable ordering: Lessons learned. In *Proceedings of the International Conference on Computer Aided Design*, pp. 646–652, San Jose, CA. IEEE/ACM.
349